\documentclass[10pt,journal]{IEEEtran}
\usepackage[]{amssymb,amsmath, graphicx, subfigure}
\begin{document}
\def\IR{{\rm I \kern-0.20em R}}
\def\bbbr{{\IR}}
\newtheorem{thm}{Theorem}
\newtheorem{prop}{Proposition}
\newtheorem{cor}{Corollary}
\newtheorem{rem}{Remark}
\newtheorem{defn}{Definition}
\newtheorem{lem}{Lemma}
\newtheorem{conjecture}{Conjecture}
\newtheorem{fact}{Fact}
\newtheorem{example}{Example}
\title{Distributed Kernel Regression: An Algorithm for Training Collaboratively}

\author{J.~B.~Predd,~\IEEEmembership{Member,~IEEE,}
        S.~R.~Kulkarni,~\IEEEmembership{Fellow,~IEEE,}
        and~H.~V.~Poor,~\IEEEmembership{Fellow,~IEEE}% <-this % stops a space
\thanks{This research was supported in part by the Army Research Office under Grant
DAAD19-00-1-0466, in part by Draper Laboratory under Grant IR\&D
6002, in part by the National Science Foundation under Grants CCR-0020524 and 
CCR-0312413, and in part by the U. S. Army Pantheon Project.}
\thanks{The authors are with the Department of Electrical Engineering, Princeton University, Princeton, NJ 08540 USA (email: jpredd@princeton.edu, kulkarni@princeton.edu, poor@princeton.edu)}\vspace{-.2in}
}% avoiding spaces at the end of the author lines is not a problem with
% conference papers because we don't use \thanks or \IEEEmembership
%Homer Simpson\authorrefmark{2},
%Montgomery Scott\authorrefmark{3} and
%Eldon Tyrell\authorrefmark{4}}
%Telephone: (800) 555--1212, Fax: (888) 555--1212}
%\authorblockA{\authorrefmark{4}Tyrell Inc., 123 Replicant Street, Los Angeles, California 90210--4321}}
\maketitle

\begin{abstract}
This paper addresses the problem of distributed learning under communication constraints, motivated by distributed signal processing in wireless sensor networks and data mining with distributed databases.  After formalizing a general model for distributed learning, an algorithm for collaboratively training regularized kernel least-squares regression estimators is derived.  Noting that the algorithm can be viewed as an application of successive orthogonal projection algorithms, its convergence properties are investigated and the statistical behavior of the estimator is discussed in a simplified theoretical setting.  
\vspace{-3mm}
\end{abstract}
\section{Introduction}
In this paper, we address the problem of \emph{distributed learning under communication constraints}, motivated primarily by distributed signal processing in wireless sensor networks (WSNs) and data mining with distributed databases. WSNs  are \textit{a fortiori} designed to make inferences from the environments they are sensing; however they are typically characterized by constraints on energy and bandwidth, which limit the sensors' ability to share data with each other or with a centralized fusion center.  In data mining with distributed databases, multiple agents (e.g., corporations) have access to possibly overlapping databases, and wish to collaborate to make optimal inferences; privacy or security concerns, however, may preclude them from fully sharing information.   Nonparametric methods studied within machine learning have demonstrated widespread empirical success in many centralized (i.e., communication \emph{unconstrained}) signal processing applications. Thus, in both the aforementioned applications, a natural question arises: can the power of machine learning methods be tapped for nonparametric inference in distributed learning under communication constraints?

In this paper, we address this question by formalizing a general model for distributed learning, and then deriving a distributed algorithm for collaborative training in regularized kernel least-squares regression.  The algorithm can be viewed as an instantiation of successive orthogonal projection algorithms, and thus, insight into the statistical behavior of these algorithms can be gleaned from standard analyses in mathematical programming.

\subsection{Related Work}
Distributed learning has been addressed in a variety of other works. Reference \cite{KeaSeu95} considered a PAC-like model for learning with many individually trained hypotheses in a distribution-specific learning framework.  Reference \cite{NguWaiJor04} considered the classical model for decentralized detection \cite{Var96} in a nonparametric setting.  Reference \cite{PreKulPoo05a} studied the existence of consistent estimators in several models for distributed learning.   From a data mining perspective, \cite{GamKegAim05} and \cite{LazObr01} derived algorithms for distributed boosting.  Most similar to the research presented here, \cite{GueBodThiPasMad04} presented a general framework for distributed linear regression motivated by WSNs.  

Ongoing research in the machine learning community seeks to design statistically sound learning algorithms that scale to large data sets (e.g., \cite{BorErtWesBot05} and references therein).  One approach is to decompose the database into smaller ``chunks", and subsequently parallelize the learning process by assigning distinct processors/agents to each of the chunks.  In principle, algorithms for parallelizing learning may be useful for distributed learning, and vice-versa. To our knowledge, there has not been an attempt to parallelize reproducing kernel methods using the approach outlined below.

A related area of research lies in the study of ensemble methods in machine learning; examples of these techniques include bagging, boosting, and mixtures of experts (e.g., \cite{FreSch97b} and others).  Typically, the focus of these works is on the statistical and algorithmic advantages of learning with an ensemble and not on the problem of learning under communication constraints.  To our knowledge, the methods derived here have not been derived in this related context, though future work in distributed learning may benefit from the many insights gleaned from this important area.

Those familiar with the online learning framework may find our collaborative training algorithm reminiscent of the equations for additive gradient updates \cite{KivWar97}.  Though both algorithms may be interpreted in the context of successive orthogonal projection algorithms, it does not appear possible to specialize the current model for distributed learning in a way that recovers the online learning framework (or vice versa).

The research presented here generalizes the model and algorithm discussed in \cite{PreKulPoo05c}, which focused exclusively on the WSN application. Distinctions between the current and former work are discussed in more detail below.

\subsection{Organization}
The remainder of this paper is organized as follows.  In Section II, we review preliminary background information necessary for the remainder of the work.  In Section III, we describe a general model for distributed learning and propose a distributed algorithm for collaboratively training regularized kernel least-squares regression estimators.  Subsequently, we analyze the algorithm's convergence properties and use these properties to gain insight into the statistical behavior of the estimator in a simplified setting. We conclude with a discussion of the method in Section IV.
\section{Preliminaries}
In this section, we briefly review the supervised learning model for nonparametric least-squares regression, reproducing kernel methods, and alternating projection algorithms.  Since a thorough introduction to these models and methods is beyond the scope of this paper, we refer the reader to standard references on the topics; see, for example, \cite{CenZen97}, \cite{GyoKohKrzWal02}, \cite{SchSmo02} and references therein.
\subsection{Nonparametric Least-squares Regression}
Let $X$ and $Y$ be ${\cal{X}}$ and ${\cal{Y}}$-valued random variables, respectively.  ${\cal{X}}$ is known as the feature, input, or observation space; ${\cal{Y}}$ is known as the label, output, or target space.  For now, we allow ${\mathcal{X}}$ to be arbitrary, but  take ${\cal{Y}}= \IR$.   In the least-squares estimation problem, we seek a decision rule mapping inputs to outputs that minimizes the expected squared error.  In particular, we seek a function $g:{\cal{X}}\rightarrow {\cal{Y}}$  that minimizes
\begin{equation}\nonumber
{\mathbf{E}}\{|g(X) - Y|^2\}.
\end{equation}
It is well-known that $\eta(x) = {\mathbf{E}}\{Y\,|X=x\}$ is the loss minimizing rule.  However, without prior knowledge of the joint distribution of $(X,Y)$, this regression function cannot be computed.  In the supervised learning model, one is instead provided a database $S=\{(x_i, y_i)\}_{i=1}^n$ of training examples with $(x_i, y_i)\in{\cal{X}}\times{\cal{Y}}$ $\forall i\in\{1,\ldots,n\}$; the learning task is to use $S$ to estimate $\eta(x)$.  

\subsection{Regularized Kernel Methods}
Regularized kernel methods \cite{SchSmo02} offer one approach to nonparametric regression.  In particular, let ${\cal{H}}_K$ denote the \emph{reproducing kernel Hilbert space} (RKHS) induced by a \emph{positive semi-definite kernel} $K(\cdot, \cdot):{\cal{X}}\times{\cal{X}}\rightarrow\IR$; let $\|\cdot\|_{{\cal{H}}_K}$ denote the norm associated with ${\cal{H}}_K$.  In practice, the kernel $K$ is a design parameter, chosen as a similarity measure between inputs to reflect prior application-specific domain knowledge.  The regularized kernel least-squares estimate is defined as the solution $f_{\lambda}\in{\cal{H}}_K$ of the following optimization problem:
\begin{eqnarray}
\label{kernel}\min_{f\in{\cal{H}}_K}\Big{[} \sum_{i=1}^n (f(x_i) - y_i)^2  + \lambda \| f \|_{{\cal{H}}_K}^2\Big{]}.
\end{eqnarray}

The statistical behavior of this estimator is well-understood under various assumptions on the stochastic process that generates the examples $\{(x_i, y_i)\}_{i=1}^n$ \cite{SchSmo02, Wah90}.  In this paper, we focus primarily on algorithmic aspects of computing a solution to (\ref{kernel})  (or an approximation thereof) in distributed environments.  To this end, consider the following ``Representer Theorem" proved originally in \cite{KimWah71}. 
\begin{thm}[\cite{KimWah71}]
Let $f_{\lambda} \in{\cal{H}}_K$ be the minimizer of (\ref{kernel}).  Then, there exists ${\mathbf{c}}_{\lambda}\in\IR^n$ such that
\begin{equation}
\nonumber f_{\lambda}(\cdot) = \sum_{i=1}^n c_{\lambda,i} K(\cdot, x_i).
\end{equation}
\end{thm}
From a computational perspective, the result is significant because it states that while the objective function (\ref{kernel}) is defined over a potentially infinite dimensional Hilbert space, its minimizer must lie in a finite dimensional subspace.  

Finally, note that (\ref{kernel}) can be naturally interpreted as an orthogonal projection.  In particular, by introducing an auxiliary vector ${\mathbf{z}}\in\IR^n$, (\ref{kernel}) can be rewritten as the following optimization program:
\begin{eqnarray}
\label{rlsqreg-r1}\min & \| {\mathbf{z}} - {\mathbf{y}}\|_2^2 + \lambda \| f \|_{{\cal{H}}_K}^2\\
\label{extraconstraints}{\textrm{s.t.}} & z_i = f(x_i) &\forall i\in\{1,...,n\}\\
\nonumber & {\mathbf{z}}\in{\IR^n}\\
\nonumber & f\in{\cal{H}}_K.
\end{eqnarray}
Through the constraints in (\ref{extraconstraints}), (\ref{kernel}) and  (\ref{rlsqreg-r1}) are equivalent in the following sense:  if $f_{\lambda}$ is the minimizer of (\ref{kernel}) and $({\mathbf{z}}^{\prime}, f_{\lambda}^{\prime})$ is the solution of (\ref{rlsqreg-r1}), then $f_{\lambda}^{\prime} = f_{\lambda}$.  Therefore, through (\ref{rlsqreg-r1}), we can interpret the regularized kernel least-squares estimator as a projection of the vector $({\mathbf{y}}, 0)\in\IR^n\times{\cal{H}}_K$ onto the set 
\begin{equation}\nonumber
\Big{\{}({\mathbf{z}}, f)\in\IR^n\times{\cal{H}}_K \, : \, z_i = f(x_i) \,\,\forall i\in\{1,...,n\}\Big{\}}\subset\IR^n\times{\cal{H}}_K.
\end{equation}
This simple observation will recur in the sequel.
\subsection{Alternating Projections Algorithms}
Let $\cal{X}$ be a Hilbert space with a norm denoted by $\|\cdot\|$.  Let $C_1, \ldots, C_m$ be closed convex subsets of $\cal{X}$ whose intersection $C = \cap_{i=1}^m C_i$ is nonempty.  Let $P_{C}(\hat{x})$ denote the orthogonal projection of ${\hat{x}}\in {\cal{X}}$ onto $C$, i.e.,
\begin{eqnarray}
\nonumber P_{C}( \hat{x}) \triangleq \arg \min_{x\in C} \| x - \hat{x}\|.
\end{eqnarray}
Define $P_{C_i}(\hat{x})$ analogously.  

Successive orthogonal projection (SOP) algorithms \cite{CenZen97} provide a natural way to compute $P_C(\cdot)$ given $\{P_{C_i}(\cdot)\}_{i=1}^m$.  For example, the (unrelaxed) SOP algorithm is defined as follows:
\begin{eqnarray}\label{SOP}
x_{0}:=\hat{x} & x_{n} := P_{C_{(n\mod{ m}) + 1} }(x_{n-1}).
\end{eqnarray}

In words, the algorithm successively and iteratively projects onto each of the subsets.  In the case where $C_i$ is a linear subspace for all $i\in\{1,\ldots,m\}$, this algorithm was first studied by von Neumann \cite{Von50}.  Often examined in the context of the \emph{convex feasibility problem}, SOP has been generalized in various ways \cite{CenZen97},  to address more general convex sets and non-orthogonal (e.g., Bregman) projections; accordingly, the algorithm often takes on other names (e.g., the von Neumann-Halperin algorithm, Bregman's algorithm).  Much of the behavior of this algorithm can be understood through Theorem 2; the proof of this  fundamental result can be found in \cite{BauBor96}.  

\begin{thm}
Let $\{C_i\}_{i=1}^m$ be a set of closed, convex subsets of $\cal{X}$ whose intersection $C = \cap_{i=1}^m C_i$ is nonempty.  Let $x_n$ be defined as in (\ref{SOP}). Then,  for every $x\in C$ and every $n\geq 1$,
\begin{equation}
\nonumber\|x_n -x\| \leq \|x_{n-1} - x\|.
\end{equation}
Moreoever, $\lim_{n\rightarrow\infty} x_n \in \cap_{i=1}^m C_i$.  If $C_i$ are affine for all $i\in\{1,...,m\}$, then $\lim_{n\rightarrow\infty} \|x_n - P_C(\hat{x})\| = 0$. \end{thm}

\section{Distributed Kernel Regression}
\subsection{The Model}
In contrast to the model for supervised learning reviewed in Section II, suppose that each member of  a collection of $m$ learning agents has limited access to the training database $S=\{(x_i, y_i)\}_{i=1}^n$.  In particular, assume that learning agent $i$ has access only to the training examples in subset $S_i\subseteq S$.    For convenience, we shall henceforth refer to $\{S_i\}_{i=1}^m$ as an \emph{ensemble}.

A bipartite graph is a convenient way to represent an ensemble in this model for distributed regression.  As depicted in Figure~1, nodes on the top-level of the graph represent learning agents; nodes on the bottom-level represent training examples.   An edge between a learning agent $i$ and a training sample $j$ signifies that agent $i$ has access to example $j$, i.e., $(x_j, y_j)\in  S_i$.  For now, we make no additional assumptions on the structural relationship between the agents' locally accessible training sets; for example, we do not require the ensemble $\{S_i\}_{i=1}^m$ to partition $S$, nor do we require the corresponding bipartite graph to be connected in any way.  

To be concrete, consider a few examples that illustrate special-cases of the general model depicted in Figure 1.  The standard centralized model for supervised learning can be represented by the graph in Figure 2, where each of the $m$ learning agents has access to all exemplars in the training database.  Figure 3 illustrates an ensemble where a publicly available database is available to all the learning agents, each of which retains a private training set.  In some applications, ${\mathcal{X}}$ may be endowed with a topology.  For example, in wireless sensor networks, ${\mathcal{X}=\IR^2}$ may model locations in a city; learning agents (i.e., sensors) may exist as points within ${\mathcal{X}}$, and query those examples  that are ``nearby" with respect to the underlying topology; such an ensemble is depicted in Figure 4.  

As mentioned earlier, the current model is a generalization of the the work discussed in \cite{PreKulPoo05c}.  Whereas \cite{PreKulPoo05c} focuses exclusively on the WSN application by assuming a topology on $\cal{X}$ and by modeling one agent per training observation, the present formulation allows a more general structure with multiple agents per training datum and an arbitrary input space.  %As a result, the subsequent analysis provides a deeper understanding of the distributed training algorithm and its application to collaborative inference.

\begin{figure}[htbp]
\centering
%\leavevmode
\includegraphics[width=3.4in]{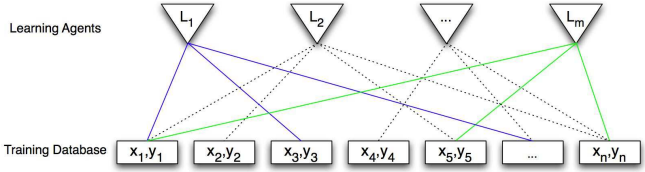}
\caption{A Bipartite Graph Representation of an Ensemble in this Model for Distributed Regression}
\label{Fig2}
\vspace{-6mm}
\end{figure}
\begin{figure}[htbp]
\centering
%\leavevmod1
\includegraphics[width=3.0in]{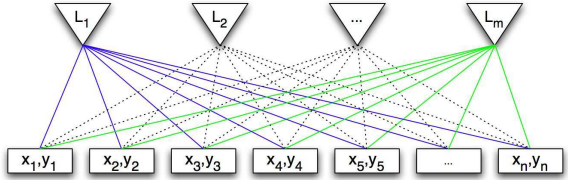}
\caption{A ``Centralized" Ensemble}
\label{Fig2}
\vspace{-6mm}
\end{figure}
\begin{figure}[htbp]
\centering
%\leavevmode
\includegraphics[width=3.0in]{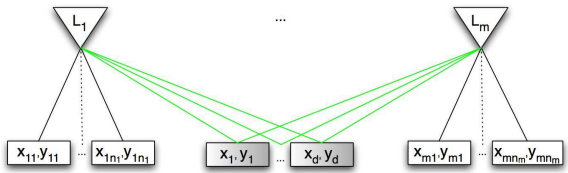}
\caption{An Ensemble with a Public Database}
\label{Fig2}
\vspace{-6mm}
\end{figure}
\begin{figure}[htbp]
\centering
%\leavevmode
\includegraphics[width=1.7in]{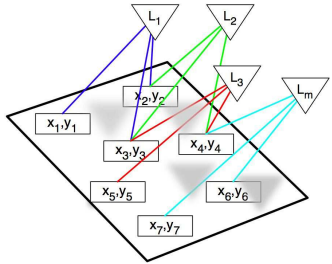}
\caption{A Sensor Network: An Ensemble with Topology Dependent Structure}
\label{Fig2}
\vspace{0mm}
\end{figure}

Presumably, each of the $m$ agents wishes to use nonparametric methods to estimate the regression function.  One simple approach is for agent $i$ to compute $f_{\lambda_i}$ using only the exemplars in its local training database $S_i$.  However, doing so ignores the structure of distributed regression and fails to exploit an opportunity to collaborate using the (partially) shared training database.  

We henceforth assume that after locally computing $f_{\lambda_i}\in{\mathcal{H}}_K$, agent $i$ may share $f_{\lambda_i}(x_j)\in\IR$ with any agent $k$ such that $(x_j, y_j)\in S_k$. In other words, neighboring agents (with respect to the bi-partite graph) communicate point estimates for the training data they share. Using such limited communication, can the agents collaborate to jointly improve the accuracy of their estimates?

%Suppose that agent $i$ can \emph{read} each example in $S_i$, and may \emph{overwrite} the label $y_j$ for any example $({\mathbf{x}}_j, y_j)\in S_i$.  In this model, the training database acts as a channel over which the agents may communicate.  Can the agents collaborate over this limited communication medium to jointly improve the accuracy of their estimates?

In the next section, we derive a collaborative training algorithm in this model for distributed nonparametric regression.   The algorithm is derived as an application of SOP algorithms applied to a relaxation of the classical regularized kernel least-squares estimator.  Subsequently, we analyze its convergence properties and investigate its statistical properties in a simplified theoretical setting.
\vspace{-2mm}
\subsection{A Collaborative Training Algorithm}
For technical convenience, let us introduce sets $\{\bar{S}_i\}_{i=1}^m$, such that $\bar{S}_i\subseteq \{1,\ldots,n\}$. Let  $j\in\bar{S}_i$ if and only if $(x_j, y_j)\in S_i$. In other words $\bar{S}_i$ contains the indices of the training examples in $S_i$ as enumerated in $S$.   Analogously, let $\bar{S} = \{1,\ldots,n\}$. 

To begin, let us rewrite (\ref{kernel}) in a way that reveals the structure of distributed regression.  To do so, first let us introduce a function $f_i\in{\mathcal{H}}_K$ for each agent $i\in\{1,\ldots, m\}$, and consider the following constrained optimization program:
\begin{eqnarray}
\label{rlsqreg-r2}\min &  \|{\mathbf{z}} - {\mathbf{y}}\|_2^2 + \sum_{i=1}^m \lambda_i \| f_i \|_{{\mathcal{H}}_K}^2\\
\label{coupling1}\textrm{s.t.} &\hspace{-1cm} z_j =  f_i(x_j) &\hspace{-1.6cm}\forall j\in \bar{S}, i\in\{1,...,m\}\\
\nonumber & \hspace{-1cm}f_i \in {\mathcal{H}}_K  &\hspace{-1.6cm}i\in\{1,\ldots,m\}
\end{eqnarray}

Here, the optimization variables are ${\mathbf{z}}\in\IR^n$ and $\{f_i\}_{i=1}^n\subset {{\cal{H}}_K}$; $S=\{(x_i, y_i)\}_{i=1}^n$ and $\{\lambda_i\}_{i=1}^m\subset\IR$ are the program data.  The \emph{coupling constraints} in (\ref{coupling1}) dictate that for any feasible solution to (\ref{rlsqreg-r2}), every agent's associated function is equivalent when evaluated at $\{x_i\}_{i=1}^n$.    As a result, one can think about (\ref{rlsqreg-r2}) as an equivalent form of (\ref{kernel}) in the following sense.

\begin{lem}
Let $({\mathbf{z}}, f_{\lambda_1},...,f_{\lambda_m})\in\IR^n\times{{\cal{H}}^m_K}$ denote the solution of (\ref{rlsqreg-r2}) and let $f_{\lambda}\in{{\cal{H}}_K}$ denote the solution of (\ref{kernel}).  Assume that $\lambda_i>0\,\,\forall i\in\{1,...,m\}$. Then, $f_{\lambda_1} = \cdots = f_{\lambda_m}$.  If $\sum_{i=1}^m \lambda_i = \lambda$, then $f_{\lambda} = f_{\lambda_1}$.
\end{lem}

This form of the regularized least-squares regression problem suggests a natural relaxation that allows us to incorporate the structure of the distributed regression model into the estimator.  In particular, we relax the coupling constraints to require that agents agree only on training examples they share:
\begin{eqnarray}
\label{rlsqreg-r3}\min &  \|{\mathbf{z}} - {\mathbf{y}}\|_2^2 + \sum_{i=1}^m \lambda_i \| f_i \|_{{\mathcal{H}}_K}^2\\
\label{coupling2}\textrm{s.t.} &\hspace{-1cm} z_j =  f_i(x_j) &\hspace{-1.6cm}\forall j\in \bar{S}_i, i\in\{1,...,m\}\\
\nonumber & \hspace{-1cm}f_i \in {\mathcal{H}}_K  &\hspace{-1.6cm}i\in\{1,\ldots,m\}
\end{eqnarray}
Thus, for any feasible solution to (\ref{rlsqreg-r3}), $f_i({\mathbf{x}}_j) = f_k({\mathbf{x}}_j)$ if $(x_j, y_j)\in S_i\cap S_k$.   Looked at in this way, (\ref{rlsqreg-r2}) models the ``centralized ensemble" depicted in Figure 2, while (\ref{rlsqreg-r3}) captures the more general structure in Figure 1.

Note that just as (\ref{kernel}) can be interpreted as a projection via (\ref{rlsqreg-r1}), (\ref{rlsqreg-r3}) can be interpreted as a (weighted) projection of the vector $({\mathbf{y}}, 0, \ldots, 0)\in\IR^n\times{{\cal{H}}^m_K}$ onto the set $C = \cap_{i=1}^m C_i$, with
\begin{eqnarray}
C_i =  \Big{\{}({\mathbf{z}}, f_1, \ldots, f_m)\,:\,\ f_i({\mathbf{x}}_j) = z_j \,\,\forall j\in \bar{S}_i, {\mathbf{z}}\in\IR^n,\\
\nonumber\,\,\,\,\,\,\,\,\,\,\,\,\,\,\,\,\,\,\, \{f_i\}_{i=1}^m\subset{{\cal{H}}_K }\Big{\}}\subset\IR^n\times{{\cal{H}}^m_K}.
\end{eqnarray}
The significance of this observation lies in the fact that the relaxed form of the regularized kernel least-squares estimator has been expressed as a projection onto the intersection of a collection of $m$ convex sets; in particular, note that each set $C_i$ is a subspace.  Thus, by Lemma 1, the SOP algorithm can be used to solve the relaxed problem (\ref{rlsqreg-r3}).  Moreover, computing $P_{C_i}(\cdot)$ requires agent $i$ to gather examples only within its locally accessible database.  More precisely, note that for any ${\mathbf{v}}=({\mathbf{z}}, f_1, \ldots, f_m)\in\IR^n\times{{\cal{H}}^m_K}$, $P_{C_i}({\mathbf{v}}) = ({\mathbf{z}}^{\star}, f_1^{\star}, \ldots, f_m^{\star})$ where
\begin{eqnarray}
\nonumber f_j^{\star} & = & f_j \,\,\,\,\,\, \forall j\neq i\\
\nonumber f^{\star}_i &=& \arg\min_{f\in{{\cal{H}}_K}} \sum_{j\in \bar{S}_i} (f(x_j) - z_j)^2 + \lambda_i \| f - f_i\|_{{\cal{H}}_K}^2\\
\nonumber z_j^{\star} & = & z_j   \,\,\,\,\,\,\forall j \textrm{ s.t. } j \notin \bar{S}_i\\
\nonumber z_j^{\star} & = & f_i^{\star}(x_j)  \,\,\,\,\,\,\forall j \textrm{ s.t. } j\in \bar{S}_i
\end{eqnarray}

To emphasize, computing $P_{C_i}({\mathbf{v}})$ leaves $z_j$  unchanged for all $j\notin \bar{S}_i$ and leaves $f_j$ unchanged for all $j\neq i$. The function associated with agent $i$, $f_{i}^{\star}$ can be computed using $f_i$ and $S_i$ after the training data labels $\{y_j\}_{j\in\bar{S}_i}$ have been \emph{updated} with the corresponding ``message variables" $\{z_j\}_{j\in\bar{S}_i}$.  Tying these observations together, we are left with an algorithm for collaborative regression estimation which solves a relaxed form of the regularized least-squares estimator (\ref{rlsqreg-r3}).  

The algorithm is summarized in psuedo-code in Table 1 and depicted pictorially in Figure 5. In words, the algorithm iterates over each agent in turn, allowing them to compute a local kernel estimate and to update the labels in the training database accordingly.  Multiple passes (in fact, $T$ cycles) over the agents are made.  
\begin{table*}[htdp]
\begin{center}
\begin{tabular}{|ll|}
\hline
\textbf{Init:} & Agents agree on a positive semi-definite kernel $K(\cdot,\cdot):{\mathcal{X}}\times{\mathcal{X}}\rightarrow\IR$.\\
& Training database $S=\{(x_i, z_i)\}_{i=1}^n$ is initialized\\
&\,\,\,\,\, so that $z_i = y_i\,\, \forall i\in\{1,\ldots, n\}$. \\
&\\
\textbf{Train:} & for $t=1,\ldots, T$ \\
&   \hspace{.5cm}for $i=1,\ldots, m$ \\
&  \hspace{1cm}Agent $i$: \\
& \hspace{1.25cm} Retrieves database $S_i\subseteq S$\\
&  \hspace{1.25cm} Computes $f_{i, t} := \arg\min_{f\in{{\cal{H}}_{K}}} \Big{[}\sum_{j\in \bar{S}_i} (f({\mathbf{x}}_j) - z_{j})^2 + \lambda_i \| f  - f_{i, t-1}\|_{{\cal{H}}_{K}}^2\Big{]}$  \\
& \hspace{1.25cm} Updates database: $z_{j} \leftarrow f_{i, t}({\mathbf{x}}_j)\,\,\,\forall (x_j, z_j)\in S_i$\\
& \hspace{.5cm} {\textrm{end}}\\
&  {\textrm{end}}\\
\hline
\end{tabular}
\end{center}
\label{default}
\caption{An Algorithm for Training Collaboratively}
\vspace{-.30in}
\end{table*}%
\begin{figure}[htbp]
\centering
%\leavevmode
\includegraphics[width=3.3in]{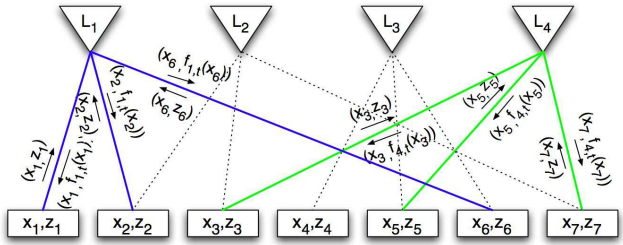}
\caption{A Collaborative Training Algorithm}
\label{Fig2}
\vspace{-4mm}
\end{figure}
\subsection{Convergence}
Note that the asymptotic behavior of the collaborative training algorithm is implied by the analysis of the SOP algorithm.  In particular,  we have the following.

\begin{thm}
Let $({\mathbf{z}}, f_{\lambda_1},\ldots, f_{\lambda_m})\in\IR^n\times{{\cal{H}}^m_K}$ be the solution to (\ref{rlsqreg-r3}) and let $\{f_{i,T}\}_{i=1}^m\subset {{\cal{H}}_K}$ be as defined in the algorithm described in Table I. Then,
\begin{equation}
\nonumber\lim_{T\rightarrow\infty} f_{i,T} = f_{\lambda_i}
\end{equation}
for all $i\in\{1,\ldots, m\}$.
\end{thm}

This theorem follows from Theorem 2 and the fact that convergence in norm implies point-wise convergence in
RKHSs.  Given the structure of RKHS and the general analysis in \cite{BauBor96}, the algorithm is expected to converge linearly for many kernels.  We forego a discussion of this important, but technical point for the sake of space.

Observe that Theorem 3 characterizes the output of collaborative training algorithm relative to (\ref{rlsqreg-r3}).  This characterization is useful insofar as it sheds light on the relationship between the algorithm's output and (\ref{kernel}), the centralized regularized least-squares estimator.  The following straightforward generalization of Theorem 1 is a step toward further understanding this important relationship.

\begin{thm}
Let $({\mathbf{z}}, f_{\lambda_1},\ldots, f_{\lambda_m})\in\IR^n\times{{\cal{H}}^m_K}$ be the solution to (\ref{rlsqreg-r3}) .  Then, for every agent $i\in\{1,\ldots,m\}$, there exists ${\mathbf{c}}_{\lambda_i}\in\IR^{|S_i|}$ such that
\begin{equation}\label{newrepresent}
f_{\lambda_i}(\cdot) = \sum_{j\in \bar{S}_i} c_{\lambda_ij} K(\cdot, x_j).
\end{equation}
\end{thm}

The proof of this theorem follows from the original  Representer Theorem (applied to the update equation for $f_{i, t}$) and the fact that ${\cal{H}}_K$ is closed.  

The significance of Theorem 4 lies in the fact that the size of any agent's locally accessible database fundamentally limits the accuracy of that agent's estimate.  In particular, an agent having access to only a few exemplars in an otherwise large training database will still be limited to estimates that lie in the span of functions determined by its local data;  thus, local connectivity influences the agent's bias.  Intuitively, however, the message-passing through the training database may optimize the estimator within that limited span if the ensemble is ``connected" in some meaningful way.  To bear out this intuition in a simplified theoretical setting, we consider a simple notion of connectedness in the next section.

\subsection{A Simplified Setting}
For a given ensemble, kernel pair $(\{S_i\}_{i=1}^m, K)$, let us construct an auxiliary graph as follows:  let there be a node for every learning agent and let there be an edge between node (i.e., agent) $i$ and node $k$  if the following condition holds:
\begin{eqnarray}
\label{connected}{\textrm{span}}(\{ K(\cdot, x_j)\}_{j\in\bar{S}_i}) &=& {\textrm{span}}( \{ K(\cdot, x_j)\}_{j\in\bar{S}_k} )\\
\nonumber&=& {\textrm{span}}(\{ K(\cdot, x_j)\}_{j\in\bar{S}_i\cap \bar{S}_k})
\end{eqnarray}
In other words, an edge connects two nodes if the training examples they share determine the space of functions their estimates lie in as dictated by Theorem 4.  

\begin{defn}
Let us call the ensemble, kernel pair $(\{S_i\}_{i=1}^m, K)$ \emph{connected} if and only if the auxiliary graph so constructed is connected. \end{defn}
This definition leads to the following theorem, which can be viewed as a straightforward generalization of Lemma 1.
\begin{thm}
Let $(\{S_i\}_{i=1}^m, K)$ be connected and suppose the ensemble employs the collaborative training algorithm using $\{\lambda_i\}_{i=1}^m$.  Finally, let $f_\lambda$ denote the solution to (\ref{kernel}) for $\lambda = \sum_{i=1}^m \lambda_i$.  Then,
\begin{eqnarray}
f_{\lambda} = \lim_{T\rightarrow\infty} f_{i, T}
\end{eqnarray}
for all $i\in\{1,\ldots, m\}$.
\end{thm}
Theorem 5 follows from Theorem 3 after noting that connectedness implies that the solution to (\ref{rlsqreg-r3}) $({\mathbf{z}}, f_{\lambda_1},\ldots, f_{\lambda_m})$ satisfies $f_{\lambda_1} =  \cdots =  f_{\lambda_m}$. To illustrate the significance of Theorem 5 and to tie it to the foregoing discussion, consider the following example. 

\begin{example}
Suppose ${\mathcal{X}}=\IR^d$ and that $K({\mathbf{x}}, {\mathbf{x}}^{\prime}) = {\mathbf{x}}^T{\mathbf{x}}^{\prime}$ is the \emph{linear kernel}; in this case, ${\mathcal{H}}_K$ is the set of linear functions on ${\mathcal{X}}$.  If $\{S_i\}_{i=1}^m$ is an ensemble with public database of $d$ \emph{linearly independent} examples (depicted in Figure 3 and discussed in Section III), then $(\{S_i\}_{i=1}^m, K)$ is connected.  Therefore, by Theorem 5, the collaborative training algorithm would allow agent $i$ to find the best linear fit to the \emph{entire data set} $S$ (for the particular choice of regularization parameter $\lambda$), despite the fact that only $\frac{n_i + d}{ \sum_{i=1}^m n_i + d}$ percent of the data is locally accessible.   More generally, if a $p^{\textrm{th}}$ order polynomial kernel is used, then an analogous observation holds when $d^p$ examples are shared.\end{example}

In this simple example, the potential utility of the collaborative training algorithm is revealed.  Consider the extreme case when each agent has access to only a single example in addition to the public database.  As the number of agents $m\rightarrow\infty$, the collaborative training algorithm would allow every agent a consistent estimate of the optimal linear least-squares estimate as long as $\sum_{i=1}^m \lambda_i \rightarrow 0$; this is true despite the fact that each agent retains local access to only $d+1$ examples for all $m$.

\vspace{-3mm}
\section{Discussion}
As described in Table 1, the inner loop of the collaborative training algorithm iterates over agents in the ensemble serially.   Note that the ordering is non-essential and parallelism may be introduced.  In fact, two agents can train simultaneously as long as they do not share exemplars in their locally accessible training database.  In practical settings, multiple-access algorithms that are frequently studied in the communications literature (e.g., ALOHA) may be adapted to negotiate an ordering in a distributed fashion.  Since the SOP algorithm and Theorem 2 have been generalized to a very general class of (perhaps random) control orderings \cite{BauBor96}, Theorem 3 can be extended in many cases.  Experiments that validate the collaborative training algorithm in a WSN setting can be found in \cite{PreKulPoo05c}. 

In this paper, we have focused exclusively on regularized kernel least-squares regression.  However using Bregman's algorithm \cite{CenZen97}, the method and many of the theorems may be extended to more general loss functions and regularizers including Bregman divergences.  %The model has also assumed that each learning agent employs the same kernel;  the model, the algorithms, and the results hold more generally in the case when the agents' kernels differ but are sufficiently similar to make (\ref{rlsqreg-r2}) feasible.

Those familiar with LDPC codes or Bayes networks may find the current model and algorithm reminiscent of message-passing algorithms such a belief-propagation which are frequently studied in those fields;  variational interpretations of kernel methods in the context of Gaussian processes further suggests a relationship between these works.  Formalizing such a connection would likely require one to interpret our ``relaxation" in the context of dependency structures in Gaussian processes, and to connect alternating projection algorithms with the generalized distributive law \cite{AjiMce00}.

\vspace{-4mm}
\bibliography{../../../../Aggregation}
\bibliographystyle{IEEEtranS}

\end{document}